# Quantifying the Effectiveness of Student Organization Activities using Natural Language Processing

[1]Lyberius Ennio F. Taruc, [2]Arvin R. De La Cruz

**Abstract**: Student extracurricular activities play an important role in enriching the students' educational experiences. With the increasing popularity of Machine Learning and Natural Language Processing, it becomes a logical step that incorporating ML-NLP in improving extracurricular activities is a potential focus of study in Artificial Intelligence (AI). This research study aims to develop a machine learning workflow that will quantify the effectiveness of student-organized activities based on student emotional responses using sentiment analysis. The study uses the Bidirectional Encoder Representations from Transformers (BERT) Large Language Model (LLM) called via the pysentimiento toolkit, as a Transformer pipeline in Hugging Face. A sample data set from Organization C, a Recognized Student Organization (RSO) of a higher educational institute in the Philippines, College X, was used to develop the workflow. The workflow consisted of data preprocessing, key feature selection, LLM feature processing, and score aggregation, resulting in an Event Score for each data set. The results show that the BERT LLM can also be used effectively in analyzing sentiment beyond product reviews and post comments. For the student affairs offices of educational institutions, this study can provide a practical example of how NLP can be applied to real-world scenarios, showcasing the potential impact of data-driven decision making.

## I. INTRODUCTION

Extracurricular activities play a vital role in enriching students' educational experiences by providing opportunities for practical and reflective learning, as well as personal growth. [10] These activities, organized by campus clubs, focus groups, and student organizations, can vary widely in nature and tone and are usually targeted at students with objectives that focus on learning and entertainment. Within the higher education setting—colleges and universities—the experience of attending and even facilitating such activities is no exception, as they provide beneficial advantages to the students' capability to learn. A study conducted in Iran found that integrating research methods into extracurricular activities significantly enhanced undergraduate nursing students' ability to conduct evidence-based care research. [2]

A case can then be made that incorporating extracurricular activities into the campus lifestyle has a direct impact on the students' emotional well-being and overall satisfaction. Extracurricular activities provide opportunities for students to develop leadership skills, build social connections, and explore their interests outside of the classroom. Moreover, engaging in such activities can provide students with opportunities to increase exposure to different skills and talents, develop friendships, and foster school connectedness [11]. Participation in various activities may also support positive school climates, with different activities supporting specific dimensions of school climate more than others [11], and can lead to positive outcomes such as motivation to learn, decreased absenteeism, lower levels of aggression, and lower suspension rates [11][12][13].

## II. RESEARCH BACKGROUND

This section explains the value of extracurricular activities, identifies the potential research gap, and describes the current state of machine learning-based natural language processing as a proposed solution.

A.*The Value of Extracurricular Activities and the Role of Student Affairs*

Extracurricular activities are generally not part of the regular school curriculum, are usually not graded, do not offer credit, and do not take place during classroom time. They are, however, still valuable to students because they provide opportunities to extend their academic and non-academic experiences under school auspices [12]. It also helps students develop important soft skills—teamwork, leadership, and time management— that, while they may seem implicit and qualitative, are highly sought after by employers in the workforce when they graduate. [13] In other words, extracurricular activities provide avenues for developing student-leadership skills. [13]

On the topic of student-organized activities, the students dedicate substantial time and resources to meticulously plan and execute their activities. The student affairs offices of schools and campuses are typically responsible for overseeing these various aspects of student life outside of the classroom setting. Typically, students or student organizations organize and sponsor these events, which offer recreation, instruction, and exercise without awarding academic credit.

[1]*Polytechnic University of the Philippines-Graduate School, FEU Institute of Technology-Computer Engineering Department,*
[2]*Polytechnic University of the Philippines-Graduate School*
[1]*lyberiusennioftaruc@iskolarngbayan.pup.edu.ph,*
*lftaruc@feutech.edu.ph,*

B. *The Problem and its Background*

Student activities are considered an integral part of the educational program of the institution [14]. However, the monitoring of student-organized extracurricular activities, specifically, is not explicitly quantified or mentioned. Institutions may still have guidelines and expectations for these activities, and they may even require student organizations to prepare event plans that include event objectives, timelines, and estimated costs, among others. In general, however, academic analytics tools are usually used to predict student success and provide proactive intervention [15] [16] for ongoing and future school events.

The lack of metrics and evaluation techniques is a common problem that student organizations and student affairs offices face when attempting to assess how effective and engaging student-organized activities are. Challenges such as lack of available resources, time constraints, and biased activity overseers may impact the accuracy and quality of these activities [17], ultimately hindering the ability to accurately assess their success.

There are existing approaches and methodologies that can be used to quantify effectiveness. One of the most common tools used in academia in general [19] is the n-point Likert scale, delivered as a feedback form to the event attendees post-activity. It can provide valuable insights into areas for improvement and overall satisfaction with the event, but should be used in conjunction with other methods to ensure a comprehensive evaluation, given that it provides data from the attendees' perspective only.

One proposed solution is to develop a machine learning-based workflow that can automate data collection and analysis processes, allowing for more efficient and unbiased evaluation of activity success. [18] Data from the event organizers can be gathered and analyzed, aside from the data collected from the attendees. This ensures that both perspectives are considered. While it can be tempting to create and issue a Likert scale-like feedback form to the organizers themselves, their biases to issue "favorable" or "positive" feedback can skew the results. Hence, a different approach should be explored.

C. *On Sentiment Analysis with Machine Learning and Python*

One approach is to use Sentiment Analysis or Opinion Mining[1], which is a technique that involves analyzing written feedback or comments to determine the sentiment or opinion of the respondent. It provides more insight by analyzing open-ended comments and classifying feedback as positive, negative, or neutral, which is not possible with Likert-based scores. Sentiment analysis, along with word clouds, can offer unique insights that go beyond predetermined Likert scale questions [28].

Sentiment analysis has become a popular research area with the rise of user-generated content on the internet. Various techniques have been developed to extract sentiments from text data, including machine learning and lexicon-based approaches. The science has gained more interest in recent years, with a significant increase in related research papers and studies, with more focus on internet-generated text from user comments, product reviews, feedback, and many others [21]. Different levels of sentiment analysis, including document level, sentence level, and word level, have been explored to improve the accuracy and efficiency of sentiment extraction [20]. Machine learning algorithms—supervised, unsupervised, and semi-supervised learning—are commonly used to classify sentiments in text data [20]. Benchmark datasets and tools, such as those from Kaggle, Meta (formerly Facebook) [25] and X (formerly Twitter) [26], are utilized for sentiment analysis tasks [20]. They are even used as dataset banks and places for data mining, as they are easily accessible by most people.

The Python programming language has also become a popular choice for sentiment analysis due to its simplicity and versatility. It has enabled advancements in Data Science and Machine Learning by providing a powerful and flexible programming language that is popular among data scientists, data engineers, enthusiasts, and hobbyists. It offers various machine learning libraries like NLTK (Natural Language Toolkit), Scikit-learn, TensorFlow, Keras, PyTorch, and Hugging Face which offer tools and algorithms to simplify the implementation of machine learning-based natural language processing workflows [27]. Overall, sentiment analysis has seen advancements in accuracy and efficiency, despite the ongoing challenges and potential improvements in the field [22] [23] [24] .

Case in point: implementing a machine learning-based sentiment analysis workflow using the data from event reports, i.e., post-activity reports, from student-organized activities is an opportunity worth exploring. While not a complete replacement for the Likert scale or other evaluation tools, it could provide a more nuanced understanding of participant experiences and feedback from a holistic perspective.

D. *Problem Statement*

Given the value student activities have for both organizers and attendees, the gap in evaluating them effectively using existing tools and methods, and the current state of machine learning and sentiment analysis, the following research problem statement can then be formulated: "What machine learning workflow incorporating sentiment analysis can be developed to quantify the effectiveness of student-organized activities based on student emotional responses?"

This research will focus on answering this question. The findings can be used as a general implementation framework for student affairs offices and even student organizations that are looking for an alternative to assess the effectiveness of student activities.

## III. RESERCH OVERVIEW

This section explains the details of the research study, including its objectives, methodology, scope, and limitations.

A. *Research Objective*

The main objective of this research is to develop a machine learning workflow that will quantify the effectiveness of student activities using sentiment analysis.

The specific objectives of this research are as follows:

• Identify the language model used for sentiment analysis;

• Identify the key features that the language model will use to generate a sentiment score for each key feature;

• Design a workflow that will generate an Event Score that considers the sentiment score of each feature and will serve as an activity's quantified score, and,

• Perform testing using sample data from an actual student organization of an educational institution, university, or college.

B. *Significance of the Study*

For students and student organizations, the study may help them better understand the effectiveness of their activities and identify which are most popular and engaging. This information can also be used to improve the quality and effectiveness of upcoming and future activities.

For the research proponents, this study can serve as a venue to apply the learnings acquired in Natural Language Processing, practice utilizing ML models, and implement the machine learning lifecycle in general.

For data scientists and enthusiasts, this study can provide a practical example of how NLP can be applied to real-world scenarios, showcasing the potential impact of data-driven decision making. Additionally, it may inspire further exploration into the field of sentiment analysis and its applications in various industries.

For the student affairs offices of educational institutions, this study can be used as a supplemental or alternative tool in evaluating how effectively their student organizations are able to deliver their activities.

For academia, this study contributes to the growing body of knowledge on the practical applications of NLP in different contexts. It also highlights the importance of staying updated with advancements in data analytics and their relevance in today's data-driven world.

C. *Methodology*

The research study covers two steps: a review of related literature and the development of a machine learning workflow using pre-trained language models.

The literature review will help to identify existing studies and papers on sentiment analysis in relation to student activities, as well as provide a foundation for understanding the current landscape of NLP in academia. It will be conducted using a systematic approach that will ensure that all the relevant sources are considered.

The development and testing of the workflow will consist of the following steps:

• **Data Collection, Generation, and Preprocessing:** Data will be collected and synthesized to become the dataset. It will also be preprocessed to remove noise and outliers. Key features are also identified, which will be used as an input to the sentiment analyzer model.

• **Model Instantiation:** A pre-trained Large Language Model (LLM) for Sentiment Analysis is instantiated. The LLM is open source and can be invoked using Python. It is given the data set features and should produce a sentiment score.

• **Feature Score Aggregation:** Each sentiment score is given a weight before being combined together to form an Event Score. The weights are arbitrary and may be changed according to importance.

• **Event Score Calculation and Generation:** Each data point is given an Event score.

• **Dataset Ranking based on Event Score:** Identify which data points have the highest event score.

D. *Sample Population and Data Set Characteristics*

*1) Confidentiality Disclaimer*

The owners granted their permission to utilize their data for the study. To ensure confidentiality and safeguard the privacy of both the students and the school involved in the study, detailed information regarding the sample population and data set characteristics has been withheld and will not be revealed. This constraint is in compliance with the Philippine Republic Act (RA) 10173, or the Data Privacy Act of 2012 (DPA) [30].

*2) Sample Population and Data Set Characteristics*

A sample dataset was derived from Organization C, one of the Recognized Student Organizations (RSO) of College X, a higher-education institution located in Manila, Philippines. It consists of individual Post-Activity reports prepared in Word document format by its officers at the time of the study. Said documents cover the events of Organization C for one academic year (SY 2022-2023), totaling twenty-one events (n=21). The nature of these events varies, from seminars to whole-day events. Organization C is an RSO for students with an interest in Computer Science, Computer Engineering, Robotics, and Mechatronics, hence, their events are geared towards the STEM field. Regardless, each data set will be treated equally.

*E. Research Scope and Limitations*

The research scope of the study is limited to analyzing the individual Post-Activity reports prepared by the officers of Organization C in Manila, Philippines, for the academic year SY 2022-2023. The data set consists of twenty-one events organized by the student officers, ranging from technical seminars to whole-day events. The study will treat each event equally in its analysis. Any personally identifiable information irrelevant to the study is removed.

## IV. A REVIEW OF RELATED LITERATURE

This section provides a review of related literature based on three key themes: 1. the general applications of Natural Language Processing and Sentiment Analysis; 2. its specific use cases applied within the context of student extracurricular activities; and, 3. an overview of *pysentimiento*: A Python Toolkit for Opinion Mining and Social NLP tasks.

*A. Applications of Natural Language Processing and Sentiment Analysis*

The applications of Natural Language Processing (NLP) have expanded significantly in recent years, with uses ranging from chatbots and language translation to governance and policymaking [7]. While its increased usage has also led governing bodies to consider regulations and ethical guidelines to ensure its responsible implementation [7], it is evident that the potential for NLP to revolutionize various industries is vast.

One approach to NLP that is of interest is Sentiment Analysis (SA), which focuses on analyzing text data to determine the sentiment or emotion expressed within the text. This sentiment consists of a Polarity, or label, e.g., positive, neutral, and negative, and an Intensity, or a numerical value representing the intensity of said sentiment, which can be a scale ranging between zero and one or between zero and one hundred [31]. This approach has led to several use cases of SA geared towards assessing and quantifying the attitude or inclination of people towards a product, service, item, etc. [8] But its use is not limited only to assessing product reviews by consumers; it may also be used by business sectors to enhance forecasting models and gain deeper insights into the market [9].

*B. Sentiment Analysis in Student Extracurricular Activities and Academia*

These papers collectively suggest that SA can be used to improve the quality of students: Iram [3] proposes an automated system to detect students' sentiments expressed in Facebook posts related to curriculum and extracurricular activities. Sirajudeen [4] presents a novel model for topic sentiment analysis in online learning communities that can provide practical references for improving the quality of information services in teaching practice. Atif [5] introduces an enhanced framework for sentiment analysis of students' surveys, aiming to improve student retention, teaching, and facilities. Gkontzis [6] proposes a data mining methodology for sentiment analysis in student forums, which can help assess the effectiveness of the learning environment and improve the overall learning experience. In general, while SA applied to student affairs and student extracurricular activities, i.e., "outside the classroom," was fairly limited at the time and relatively new of a field, there is still relevance for SA to be used within academia. Such approaches still contribute to the advancement of SA techniques in educational settings, ultimately benefiting students and members of the academic community by introducing another tool for assessing student satisfaction and engagement.

*C. Pysentimiento Overview and the BERT Large Language Model (LLM)*

Pysentimiento is an (NLP) toolkit in Python designed for opinion mining and Social NLP tasks. Built on top of the HuggingFace's Transformers library, pysentimiento offers a simple API to utilize pre-trained Language models for sentiment analysis and other social NLP tasks. By selecting datasets for specific tasks and language pairs, pre-processing them for model training, fine-tuning various underlying models, comparing their performance, and integrating the best models into the toolkit, pysentimiento provides an accessible and efficient solution for extracting opinions and information from user-generated text across multiple languages. [32]

The study uses the Bidirectional Encoder Representations from Transformers (BERT) Large Language Model (LLM) called via the pysentimiento toolkit, as a Transformer pipeline in Hugging Face. BERT is a transformer-based pre-trained language model that has significantly improved performance on various natural language processing tasks. It is trained with two learning objectives: masked language modeling (MLM) and next sentence prediction (NSP) to learn semantic information within and between sentences. BERT models are widely used in the NLP community for their effectiveness in understanding language semantics and context [33]. One of its advantages over other pre-trained LLMs is its high accuracy and performance, especially for social media data like X tweets. Although unrelated to the study, BERT LLM models can handle different languages and dialects as they are trained on multilingual data, making them suitable for analyzing sentiment in text written in various languages [34].

BERT is used in the study over other available LLMs such as LLaMA 2, FALCON, and BLOOM because not only is the language model open source, with a large community supporting its development [35], but also its general availability, after having pre-tained with large amounts of text data, including social media content [34].

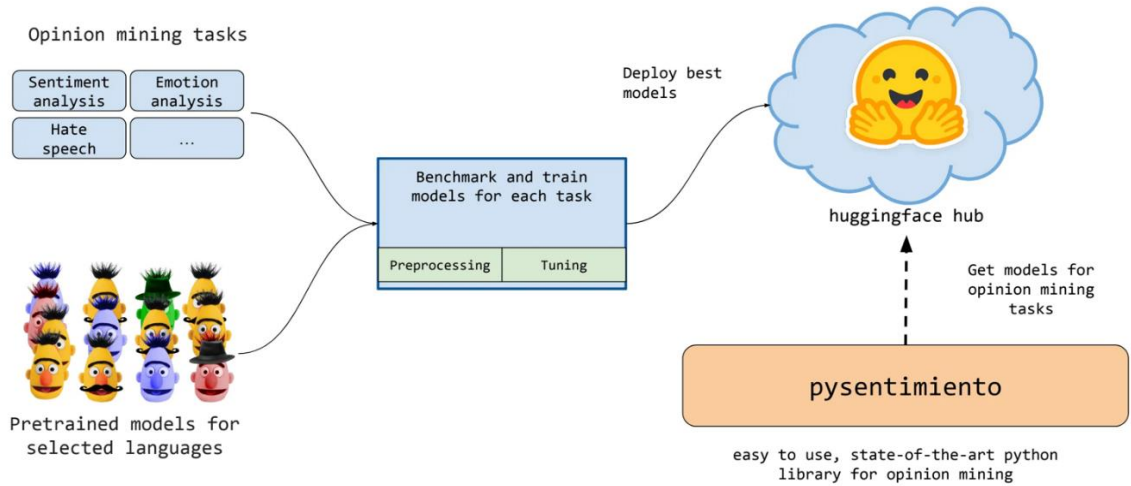

**Fig. 1.** pysentimiento High-Level Architecture.

## V. WORKFLOW OVERVIEW, AND DATA & RESULTS OVERVIEW

This section explains the designed workflow using the Hugging Face Transformer pipeline, the key features used in Sentiment Analysis (SA), and the results of NLP-SA using the sample dataset.

*A. Features Selection*

Going through the sample data, the following is a summary of the identified variables and a description of each, *Table I*:

*Table I. Sample Data Variables.*

| Parameter Name | Assigned Variable | Data Type | Description and Characteristics | Sample values |
|---|---|---|---|---|
| Activity Name | N | String | Name of the proposed activity | *"Neebot Presents CpESemfields: Living in the Cloud"* |
| Date of Activity | D | Date | Date when the activity was conducted, in MM/DD/YYYY format | *10/15/2022* |
| Duration | U | Integer | Duration of the activity in days | *1* |
| SY | S | String | Academic School year of which the activity was conducted | *2022-2023* |
| Term | T | String | School Semester | *1T (which stands for the First Term of the school year)* |
| Problems Encountered | P | String | Descriptive overview of the problems encountered by the organizers before, during and after the activity | *"…Conflict of schedule because there is a midterm examination during the time allotted for the seminar…"* |

| | | | | | |
|---|---|---|---|---|---|
| Recommendations | R | String | The recommendations proposed by the organizers to improve the event should the organization wish to conduct the activity again in the future. | *"… if you reserved online there is no student discount available. You need to decide whether to secure seats online with no student discounts or go directly at the bus station…"* |
| Conclusion | C | String | Descriptive summary of the event, whether it was successful or not based on the event's proposed objectives. | *"…This event is still a success despite its delayed execution due to the unavailability of the officers to handle it…"* |

Assessing the nature of each variable, the following details have been identified: Activity Name (N) is used to label one data row, as this contains the Activity Name. The key features passed to the SA model are Problems Encountered (P), Recommendations (R), and Conclusion (C), because they contain the sentiments describing N. Since the objective of the study focuses on general sentiment and the timing of each N row is not relevant in assessing each row, variables D,U,S, and T are trimmed during data preprocessing.

*B. Model Instantiation using the Hugging Face Transformer Pipeline*

The BERT LLM used in the study is invoked via the Hugging Face Transformer pipeline() function, which came from the pysentimiento library for opinion mining. It is assigned as a Python object using the command:

*sentiment_pipeline =
pipeline(model="finiteautomata/bertweet-base-sentiment-analysis")*

As a pre-trained model, the default parameters were retained, and no additional fine-tuning was done for the specific task of sentiment analysis in this study.

In NLP-SA, each data set that is passed through the LLM is given a Polarity and a Score or Intensity [31]. The BERT model in the study produced the following polarity and score value ranges [36], which are stored as two-element lists in Python, *Table II*:

**Table II.** pysentimiento BERT Sentiment Analysis Values [32].

| Value Name | Characteristic | Values |
|---|---|---|
| 'label' | Polarity label: 'POS' for Positive, 'NEG' for Negative, 'NEU' for Neutral. | ['POS','NEG',NEU'] |
| 'score' | Intensity or Score label which is an eight-decimal value between negative one and positive one. | [-1..1] |

*C. Feature Score Aggregation and Event Score Generation*

Because the data set considers three key features, there is a need to aggregate them into a single value, the Event Score (Y), which is associated with each row N. Assessing the descriptive nature of each key feature, the following arbitrary score was formulated:

$$EventScore = 0.2P + 0.4R + 0.4C$$

The Event Score formula considers the weighted average of the three key features to generate a score for each row N. It will have a value range between zero and one, which can be further expressed as a percentage over 100%. While each feature can be given equal weights, Problems Encountered, P, is given a 0.2 weight because it is considered less impactful compared to the other two features, R and C. It is also expected that P will produce a generally negative sentiment, as the characteristic describes the general problems encountered by Organization C when conducting the event N.

*D. Workflow Overview*

The diagrams below illustrates the workflow for deriving Event Score Y from key features P, R, and C, *Figs. 2, 3, and 4*:

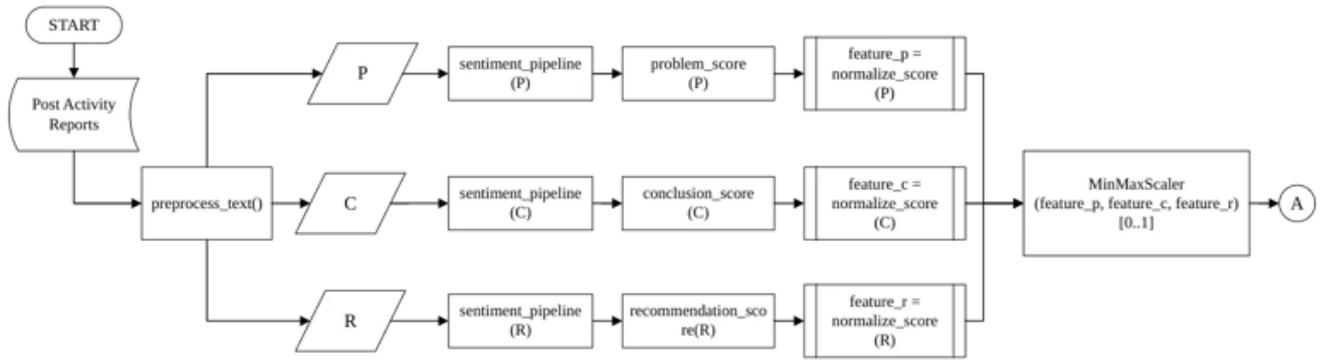

**Fig. 2**. Event Score Workflow Overview (1)

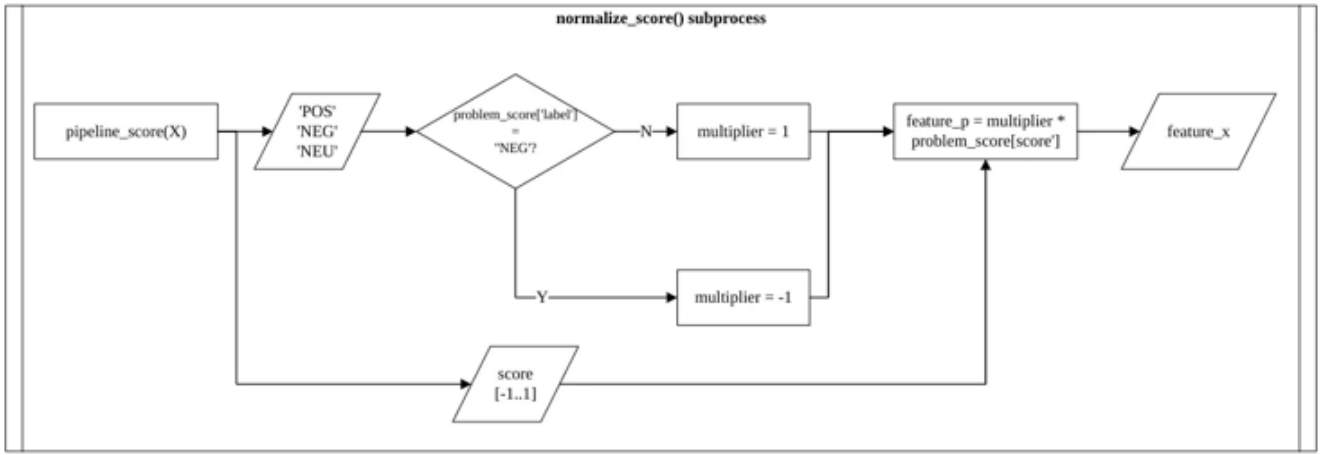

**Fig. 3**. normalize_score() Subroutine workflow.

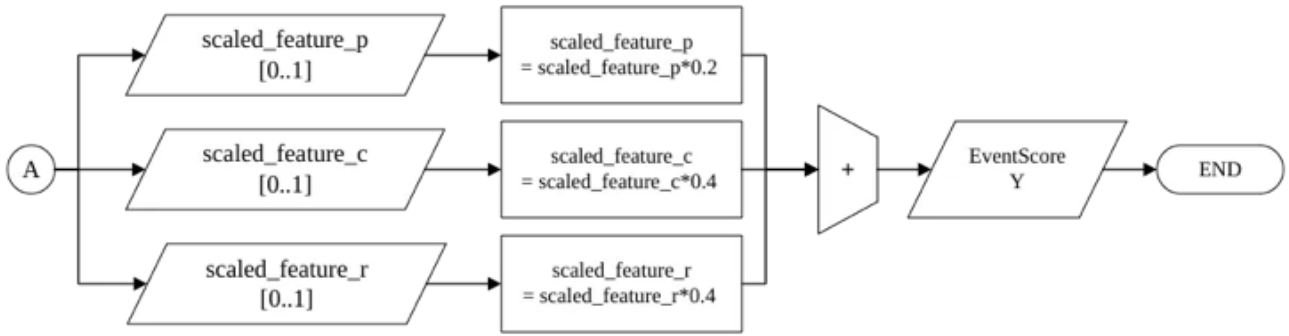

**Fig. 4.** Event Score Workflow Overview (2)

The workflow diagram outlines a process for calculating an EventScore (Y) using three key features: P, C, and R. Because the data set came from raw individual word document files, they first went through data extraction and consolidation, resulting in a single CSV file. The features are then extracted using Python's data science libraries: numpy, pandas, and sckitlearn. The result is a data frame containing the features of the study. Each feature is then routed through a sentiment pipeline using the Hugging Face Transformer library, where it passes through the pre-trained BERT LLM, instantiated via the sentiment_pipeline() object. These scores are then normalized between 0 and 1 using the normalize_score() subprocess and MinMaxScaler function. Finally, the scores are weighted (P: 0.2, C: 0.4, R: 0.4) and summed together to arrive at the final EventScore. This score is then associated to each data row N.

## VI. Data, Workflow Results, and General Findings

After passing through the data set in the workflow, the following table is created:

Table III. Event Score Workflow Results.

| | Activity Name | Normalized "Problems Encountered" Score | Normalized "Conclusions" Score | Normalized "Recommendations" Score | Event Score |
|---|---|---|---|---|---|
| 0 | Providing Security Connectivity with SD-WAN… | 3.07% | 89.36% | 84.07% | 69.99% |
| 1 | Preparing Myself Future-Ready… | 0.00% | 90.42% | 95.09% | 74.21% |
| 2 | Living in the Cloud… | 99.61% | 88.48% | 96.01% | 93.72% |
| 3 | The Future with Data… | 98.75% | 89.82% | 95.79% | 94.00% |
| 4 | The Future of Sustainable and Reliable Wireless Architecture… | 100.00% | 92.02% | 85.67% | 91.08% |
| 5 | Financial Awareness… | 98.75% | 99.55% | 94.11% | 97.22% |
| 6 | Enabling Data Analytics in Semiconductor Manufacturing Company… | 95.90% | 89.74% | 99.08% | 94.71% |
| 7 | Role of Research and Development in Industrial 4.0 | 8.04% | 96.03% | 81.03% | 72.43% |
| 8 | The Finale Battle… | 83.19% | 100.00% | 91.87% | 93.39% |
| 9 | National Challenge 2023… | 99.56% | 78.38% | 96.74% | 89.96% |
| 10 | 1st Hybrid Technical Seminar 2023 - Industrial Robot Programming… | 97.06% | 98.57% | 100.00% | 98.84% |
| 11 | KadaKareer x HelpBridge HacKada | 14.30% | 99.08% | 94.40% | 80.25% |
| 12 | Arduino Day Philippines 2023 | 97.93% | 99.18% | 0.00% | 59.26% |
| 13 | …Boot Up | 4.02% | 10.69% | 98.76% | 44.58% |
| 14 | Python Review… | 5.36% | 99.48% | 97.97% | 80.05% |
| 15 | Samsung Create… | 98.03% | 89.43% | 91.85% | 92.12% |
| 16 | Java Review… | 98.75% | 97.92% | 95.79% | 97.23% |
| 17 | The Grand Cyberleague Expo… | 1.78% | 0.00% | 78.03% | 31.57% |
| 18 | CpE Trilogy: Cybersecurity… | 8.24% | 95.23% | 91.62% | 76.39% |
| 19 | …Building… | 0.50% | 87.84% | 78.57% | 66.66% |
| 20 | Python Review | 98.75% | 98.17% | 95.79% | 97.34% |

Among the various activities, "Living in the Cloud" scored exceptionally well at 93.72%, indicating minimal problems and well-developed aspects. The "National Challenge" achieved a strong score of 89.96%, showcasing solid conclusions and recommendations despite some encountered issues, while "Industrial Robot Programming" stood out with a near-perfect score of 98.84%, demonstrating excellent performance in conclusions, recommendations, and problem-encounters. These high-scoring activities can infer the success and quality of the conducted activities by Organization C.

On the other hand, it can also be highlighted that there are also low-scoring events: the "Grand Cyberleague Expo" scored poorly across all categories with a lack of conclusions; "Boot Up" might have experienced a struggle in all areas, facing various problems and offering weak conclusions and recommendations. Similarly, "…Building…" had strong recommendations, but might have faced several issues and provided weak conclusions.

Generally, high Event Scores correlate with conclusions and recommendations that share a strong positive sentiment.

Interestingly, a high P-value does not necessarily lead to a low Event Score: "…National CpE Challenge…" has a satisfactory Event Score despite encountering some problems, likely due to its strengths in other areas.

CONCLUSION

Through the implementation of Python libraries and tools such as scikit-learn, pandas, and the Hugging Face Transformer, an EventScore for every data row was derived, thereby effectively quantifying each activity using only key features. By following this methodical approach, a full understanding of the study's sentiment has been reached, which allows for the extraction of further insights and analyses from the results.

This study has also proven that the BERT LLM can also be used effectively in analyzing sentiment beyond product reviews and post comments. In general, using these advanced tools and methods has made opinion mining, within the context of academia and student affairs, much more accurate and useful in many situations. In the future, adding more NLP-based use cases and analyzing a wider range of data can lead to even more complete and useful results.